%% file: iccad21.tex
\documentclass[conference]{IEEEtran}
\IEEEoverridecommandlockouts
\usepackage{cite}
\usepackage{amsmath,amssymb,amsfonts}
\usepackage{algorithmic}
\usepackage{graphicx}
\usepackage{textcomp}
\usepackage{xcolor}
\usepackage{threeparttable}
\def\BibTeX{{\rm B\kern-.05em{\sc i\kern-.025em b}\kern-.08em
    T\kern-.1667em\lower.7ex\hbox{E}\kern-.125emX}}
\usepackage{multirow}
\usepackage[caption=false]{subfig}
\usepackage{tabu}
\usepackage{color, colortbl}
\newcommand{\thickhline}{%
    \noalign {\ifnum 0=`}\fi \hrule height 1pt
    \futurelet \reserved@a \@xhline
}
\usepackage[linesnumbered,ruled]{algorithm2e}
\usepackage{xcolor}

\SetCommentSty{mycommfont}    
\usepackage{pifont}
\usepackage{cite}
\usepackage{amsmath}
\newtheorem{Tlemma}{Lemma}

\newtheorem{Tdef}{Definition}
\newenvironment{Definition}[1]
    {\begin{Tdef}\noindent\textsc{(#1)}\itshape}
    {\end{Tdef}}
\usepackage{amsmath}

\newcommand{\actor}[1]{\textbf{\emph{{#1}}}}
\newcommand{\ineq}[1]{\footnotesize$#1$\normalsize}{}
{}

\begin{document}
\bstctlcite{IEEEexample:BSTcontrol}
% \title{Conference Paper Title*\\
\title{Design-Technology Co-Optimization for \\NVM-based Synaptic Processing Unit}
\title{Design-Technology Co-Optimization for \\NVM-based Neuromorphic Processing Elements}
\title{\huge Dataflow-based Mapping Explorations for Performance Optimization of Neuromorphic Computing}
\title{A Design flow for Performance Estimation of Neuromorphic Computing}
%\title{A Predictable Design Flow for Mapping Spiking Neural Networks to Neuromorphic Hardware}
\title{A Design Flow for Mapping Spiking Neural Networks to Many-Core Neuromorphic Hardware}

\author{\IEEEauthorblockN{Shihao Song, M. Lakshmi Varshika, Anup Das, and Nagarajan Kandasamy}
\IEEEauthorblockA{\textit{Electrical and Computer Engineering, Drexel University} \\
%\textit{Drexel University}\\
Philadelphia, PA, USA \\
Email: \{shihao.song,lm3486,anup.das,nk78\}@drexel.edu}
% \and
% \and
% \and
% \and
% \and
}
%\author{}

\maketitle

\begin{abstract}
\input{sections/abstract}
\end{abstract}


\section{Introduction}\label{sec:introduction}
\input{sections/introduction}

\section{Background and Related Works}\label{sec:background}
\input{sections/background}

\section{Proposed Design Flow}\label{sec:design_flow}

\input{sections/designflow}

\section{Dataflow Representation of SNNs}\label{sec:sdfg_formulation}
\input{sections/sdfg}

\section{Iterative SNN Partitioning}\label{sec:iterative_cut}
\input{sections/partitioning}

\section{Hardware Mapping Explorations}\label{sec:performance_tradeoff}
\input{sections/mapping}

\section{Evaluation Methodology}\label{sec:evaluation}
\input{sections/evaluation}

\section{Results and Discussions}\label{sec:results}
\input{sections/results}

\section{Conclusions}\label{sec:conclusions}
\input{sections/conclusions}

\section*{Acknowledgement}
This work is supported by DE-SC0022014, CNS-2008167, and CCF-1937419.

%\clearpage
\bibliographystyle{IEEEtran}
\IEEEtriggeratref{32}
\bibliography{commands,disco,external}

\end{document}

%% file: sections/abstract.tex
The design of many-core neuromorphic hardware is getting more and more complex as these systems are expected to execute large machine learning models.
To deal with the design complexity, a predictable design flow is needed to guarantee real-time performance such as latency and throughput without significantly increasing the buffer requirement of computing cores. 
Synchronous Data Flow Graphs (SDFGs) are used for predictable mapping of streaming applications to multiprocessor systems.
We propose an SDFG-based design flow for mapping spiking neural networks (SNNs) to many-core neuromorphic hardware with the objective of exploring the tradeoff between throughput and buffer size. The proposed design flow integrates an iterative partitioning approach, based on Kernighan–Lin graph partitioning heuristic, creating SNN clusters such that each cluster can be mapped to a core of the hardware. The partitioning approach minimizes the inter-cluster spike communication, which improves latency on the shared interconnect of the hardware. Next, the design flow maps clusters to cores using an instance of the Particle Swarm Optimization (PSO), an evolutionary algorithm, exploring the design space of throughput and buffer size. Pareto optimal mappings are retained from the design flow, allowing system designers to select a Pareto mapping that satisfies throughput and buffer size requirements of the design. We evaluated the design flow using five large-scale convolutional neural network (CNN) models. Results demonstrate 63\% higher maximum throughput and 10\% lower buffer size requirement compared to state-of-the-art dataflow-based mapping solutions.

%% file: sections/introduction.tex
Neuromorphic computing systems are integrated circuits that implement the architecture of a central nervous system of primates~\cite{mead1990neuromorphic,bose2019my,christensen20212021}. These systems enable energy-efficient execution of Spiking Neural Networks (SNNs)~\cite{maass1997networks} due to their event-driven execution, low-power design, and distributed in-place neural computing and synaptic storage architecture. Therefore, neuromorphic systems are suitable for implementing machine-learning inference tasks on Embedded Systems and Edge devices of the Internet-of-Things.

A neuromorphic hardware is implemented as a many-core architecture, where a core is a processing element (PE) consisting of neuron circuitry and memory cells~\cite{catthoor2018very}.
%In a neuromorphic system, cores are interconnected via a shared interconnect such as Network-On-Chip (NOC)~\cite{liu2018neu} and Segmented Bus~\cite{balaji2019exploration}.
%that are interconnected using a shared interconnect such as Network-on-chip (NOC)~\cite{liu2018neu} and Segmented Bus~\cite{balaji2019exploration}. 
A common design practice is to build a PE as an analog crossbar~\cite{liu2015spiking}, where memory cells are organized in a two-dimensional grid with horizontal wordlines and vertical bitlines connecting the neuron circuitry as illustrated in Figure~\ref{fig:crossbar}. 
%where 
%memory cells storing the synaptic weights are placed at the crosspoint of bitlines and wordlines
%placed on a verical wire called bitline. 
%organized in a grid with bitlines and wordlines.
%bitlines and wordlines are organized in a grid with memory cells connected at their crosspoints to store synaptic weights.
%,hu2014memristor,hu2016dot,ankit2017trannsformer,nukala2014spintronic,kim2012digital,zhang2018neuromorphic,gopalakrishnan2020hfnet,fernando20203d}.
%~\cite{catthoor2018very}. 
%Pre- and post-synaptic neuron circuits are connected on wordlines and bitlines, respectively (see Figure~\ref{fig:crossbar}).
%To increase the energy efficiency of neuromorphic systems, Non-Volatile Memory (NVM) such as oxide-based random access memory (OxRRAM) is used to store analog synaptic weights in a crossbar~\cite{mallik2017design}.
%An NVM cell can implement multi-bit synaptic weight:
%An NVM cell can implement analog synaptic weights~\cite{shim2020impact}.

A crossbar can accommodate only a fixed number of pre-synaptic connections per post-synaptic neuron. Its dimension is typically constrained to reduce energy consumption and mitigate the negative impact of technology scaling. Therefore, neuromorphic system software frameworks 
%such as NEUTRAMS~\cite{ji2016neutrams}, NeuroXplorer~\cite{neuroxplorer}, and DFSynthsizer~\cite{dfsynthesizer} 
partition SNNs into smaller clusters such that each cluster can be mapped directly on to the crossbar of a neuromorphic PE~\cite{psopart}.
%\footnote{There are also other system software frameworks such as~\cite{loihi_mapping,zyarah2020neuromorphic,zhang2020lifetime,twisha_endurance,espine,khan2008spinnaker,reneu,song2020case,vts_das,lee2019system,twisha_thermal,corelet}.}
We show that existing frameworks are either not scalable to large problem sizes or their exploration strategies do not encompass large portions of the hardware mapping design space, leaving behind a significant opportunity to improve performance.

Typically, inference hardware platforms are expected to perform streaming machine learning tasks, i.e., to perform machine learning inference continuously on streaming data collected from different sensors. A key performance metric for such tasks is the throughput, defined and the inverse of the time it takes to perform an inference (i.e., the time between when an input is presented and when an outcome is returned by the hardware). A neuromorphic computing inference hardware enables parallel execution and pipelining of operations. Therefore, scheduling operations of a machine learning inference task onto this pipelined parallel computing environment is a grand challenge. Additionally, once a machine learning task is partitioned into clusters, cyclic dependency may exist between these clusters, which can lead to performance degradation or in the worst-case, execution deadlock. 

In this paper, we propose a design flow for mapping SNN-based machine learning applications onto the PEs of a many-core neuromorphic hardware with a predictable timing behavior. 
We make the following four key \textbf{contributions}.
\begin{itemize}
    \item We propose an iterative approach to partition an SNN into smaller clusters such that each cluster can be implemented on a PE. Our iterative approach integrates the Kernighan–Lin graph partitioning heuristic to finding a set of minimum cuts of the directed graph representation of an SNN, minimizing the data (spike) communication between clusters (see Section~\ref{sec:iterative_cut}).
    \item We exploit the rich semantics and expressiveness of Synchronous Data Flow Graphs (SDFGs) to represent SNNs, allowing us to analyze key performance properties such as throughput and buffer space, incorporating resource constraints of the hardware (See Section~\ref{sec:sdfg_formulation}).
    \item We propose a framework to analyze consistency and deadlock when mapping machine learning clusters to hardware. The framework allows to estimate the throughput degradation obtained when 1) the buffer size in each PE is limited and 2) when the PEs need to be time-multiplexed between different clusters (see Section~\ref{sec:performance_tradeoff}).
    \item We present a design flow 
    for mapping SNN-based machine learning applications to state-of-the-art many-core neuromorphic computing systems using an instance of the Particle Swarm Optimization (PSO). The mapping solutions of the PSO heuristic explores the design space of performance and buffer size (see Section~\ref{sec:design_flow}).
    
\end{itemize}

We evaluate our design flow for a recent neuromorphic hardware using convolutional neural network (CNN)-based machine learning applications. Results show the scalability of our solution and a significant improvement in throughput. 

\begin{figure*}[h!]%
    \centering
    \subfloat[An LIF neuron with $N$ pre-synaptic connections.\label{fig:lif}]{{\includegraphics[width=5.8cm]{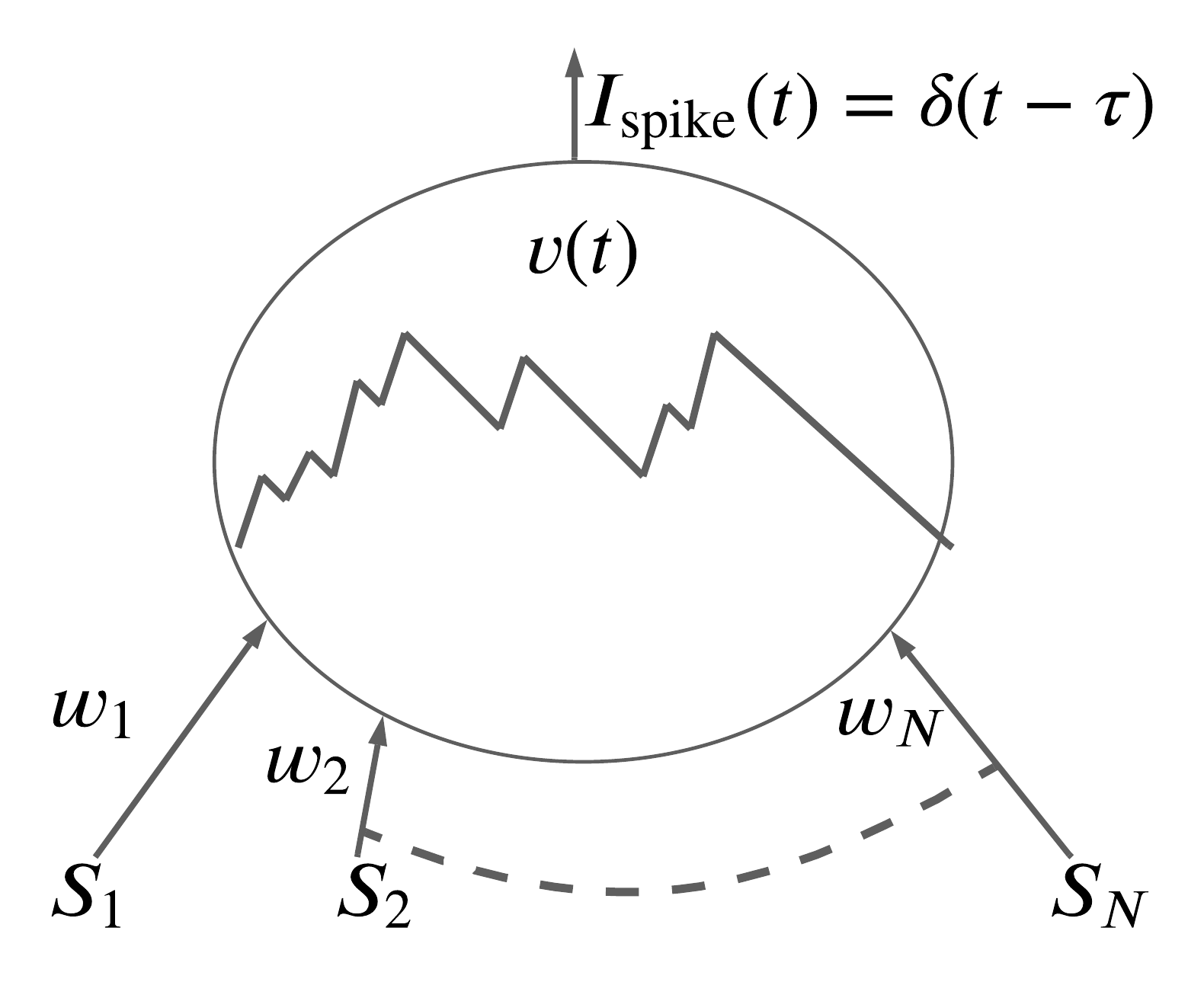} }}%
    \quad
    \subfloat[Implementation of an LIF neuron.\label{fig:implementation}]{{\includegraphics[width=4.2cm]{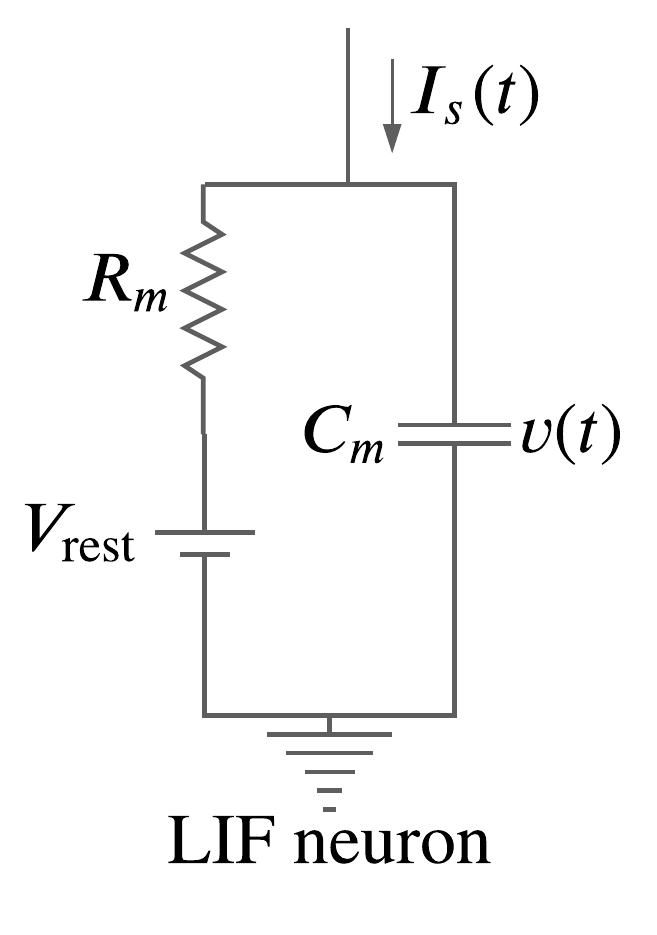} }}%
    \quad
    \subfloat[State diagram of an LIF neuron.\label{fig:state_diagram}]{{\includegraphics[width=5.0cm]{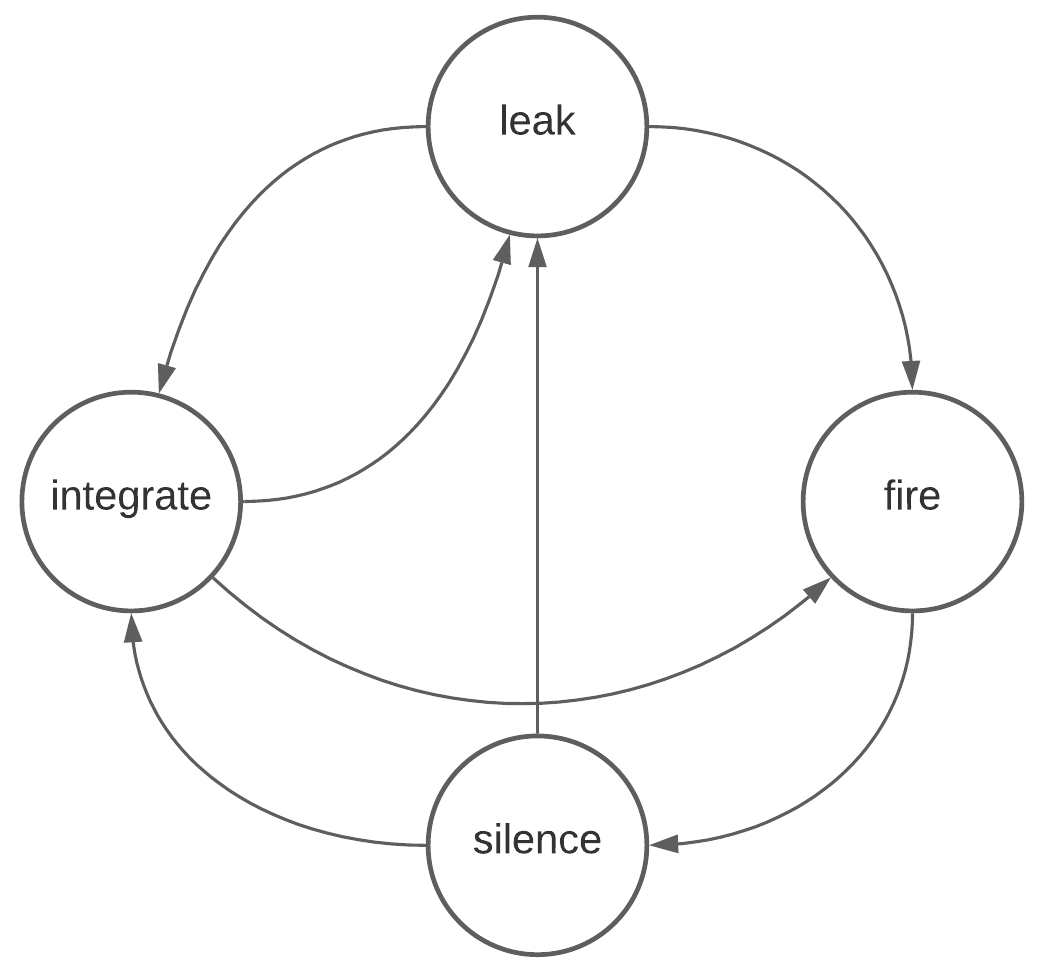} }}%
    \caption{Implementation and operation of an LIF neuron.}%
    \label{fig:arch_implementation}%
\end{figure*}

%% file: sections/background.tex
Spiking Neural Networks (SNNs) enable powerful computations due to their spatio-temporal information encoding capabilities~\cite{maass1997networks}. Figure~\ref{fig:lif} shows the operation of a leaky integrate-and-fire (LIF) post-synaptic neuron with \ineq{N} pre-synaptic connections. The neuron is described by the state variable \ineq{v(t)}, which represents the membrane potential of the neuron. Figure~\ref{fig:implementation} shows a simple implementation of the neuron using membrane resistance \ineq{R_m} and capacitance \ineq{C_m}.  

Figure~\ref{fig:state_diagram} shows the state diagram of the neuron.
%In the \emph{integrate} state, the neuron membrane potential increases due to the injected current from its pre-synaptic connections. 
The dynamics of the neuron is described by~\cite{burkitt2006review} 
\begin{equation}
    \label{eq:neuron_integrate}
    \footnotesize C_m \frac{dv(t)}{dt} = I_\text{leak}(t) + I_s(t) + I_\text{inj}(t),
\end{equation}
where \ineq{I_\text{leak}(t) = -\frac{C_m}{\tau_m}[v(t) - v_\text{rest}]} is the leakage current in the membrane, \ineq{\tau_m = C_m R_m} is the time constant of the membrane, \ineq{v_{rest}} is the resting potential, \ineq{I_s(t)} is the current due to the synaptic input to the neuron, and \ineq{I_\text{inj}(t)} is the current injected into the neuron by an intercellular electrode.

We consider current-based (CUBA) synapses, where the synaptic current of the post-synaptic neuron is given by
\begin{equation}
    \label{eq:cuba}
    \footnotesize I_s(t) = \sum_{i=1}^N S_iW_i,
\end{equation}
where \ineq{S_i = \sum_{\tau_k} \delta(t-\tau_k)} is the spike train of \ineq{i^\text{th}} pre-synaptic neuron and \ineq{w_i} is the synaptic strength of the connection of this neuron to the post-synaptic neuron.

In the firing state, the post-synaptic neuron fires a spike when its membrane voltage \ineq{v(t)} crosses the threshold voltage \ineq{V_\text{th}}. The output spiking current is defined as
\begin{equation}
    \label{eq:spiking_mechanism}
    \footnotesize I_\text{spike}(t) = C_m\left[\frac{dv(t)}{dt}\right]_{v=V_\text{th}}^{-1} (V_\text{rest} - V_\text{th})\delta(v(t)-V_\text{th})
\end{equation}

%In an SNN, spikes (i.e., current) injected from pre-synaptic neurons raises the membrane voltage of a post-synaptic neuron. When the membrane voltage crosses a threshold, the post-synaptic neuron emits a spike that propagates to other neurons. SNNs implement some variants of Integrate and Fire (I\&F) neurons with a spike duration ranging from 1 $\mu$s to several ms~\cite{ifneuron,zhang2020pulse,leigh2020efficient}. 
SNNs can implement many machine learning approaches.
For a supervised machine learning application,
%is the primary focus of this work. Here,
%relevant to this work being the supervised machine learning, where 
%a model 
an SNN is trained with representative data, where training refers to adjusting the synaptic weight of connections between pre- and post-synaptic neurons of the SNN~\cite{wu2021tandem}. Machine-learning \textbf{inference} refers to feeding live data points to a trained SNN and generating the corresponding output.

\begin{figure}[h!]%
    \centering
    \subfloat[A many-core neuromorphic hardware.\label{fig:tile}]{{\includegraphics[width=3.8cm]{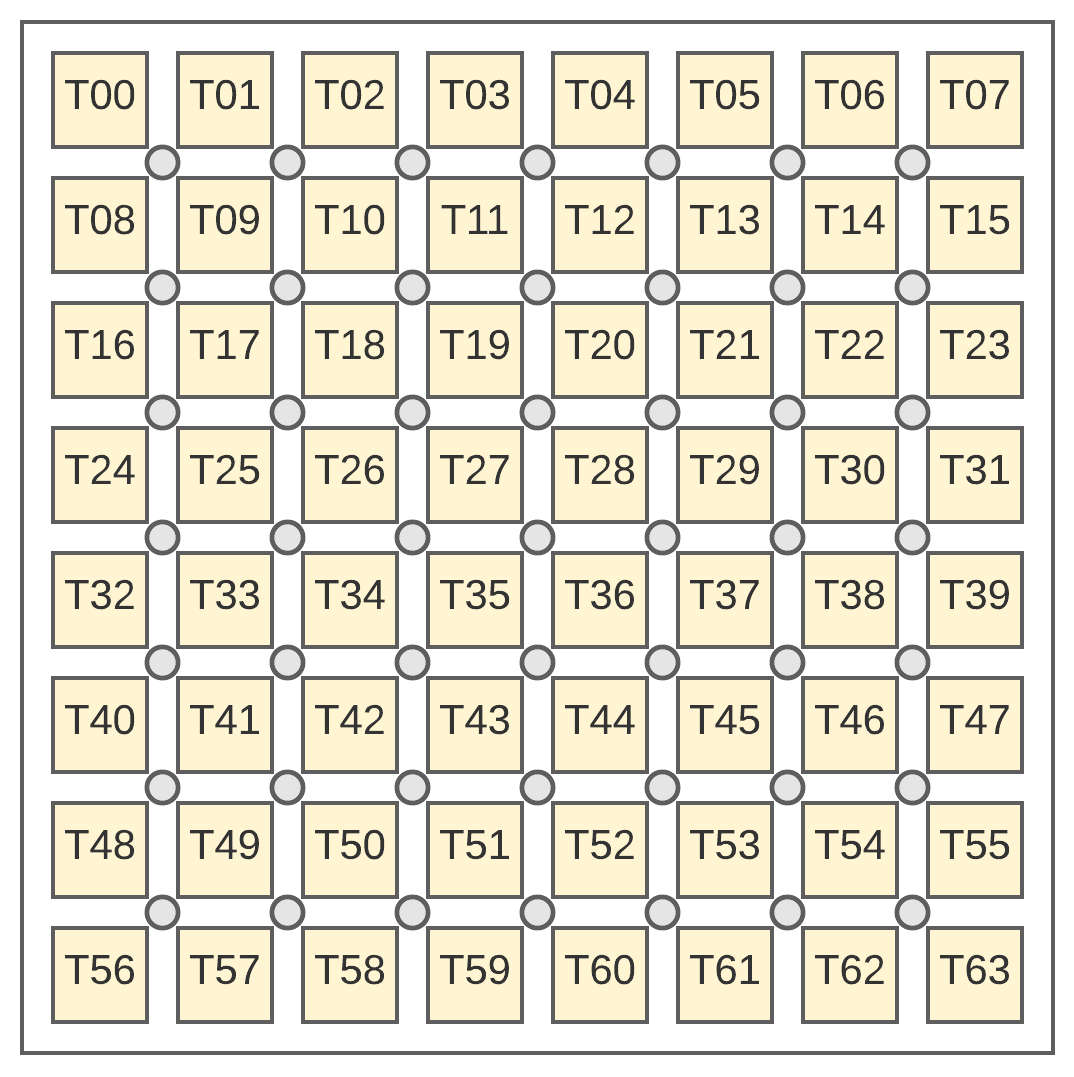} }}%
    %\quad
    \subfloat[Analog crossbar-based PE.\label{fig:crossbar}]{{\includegraphics[width=4.5cm]{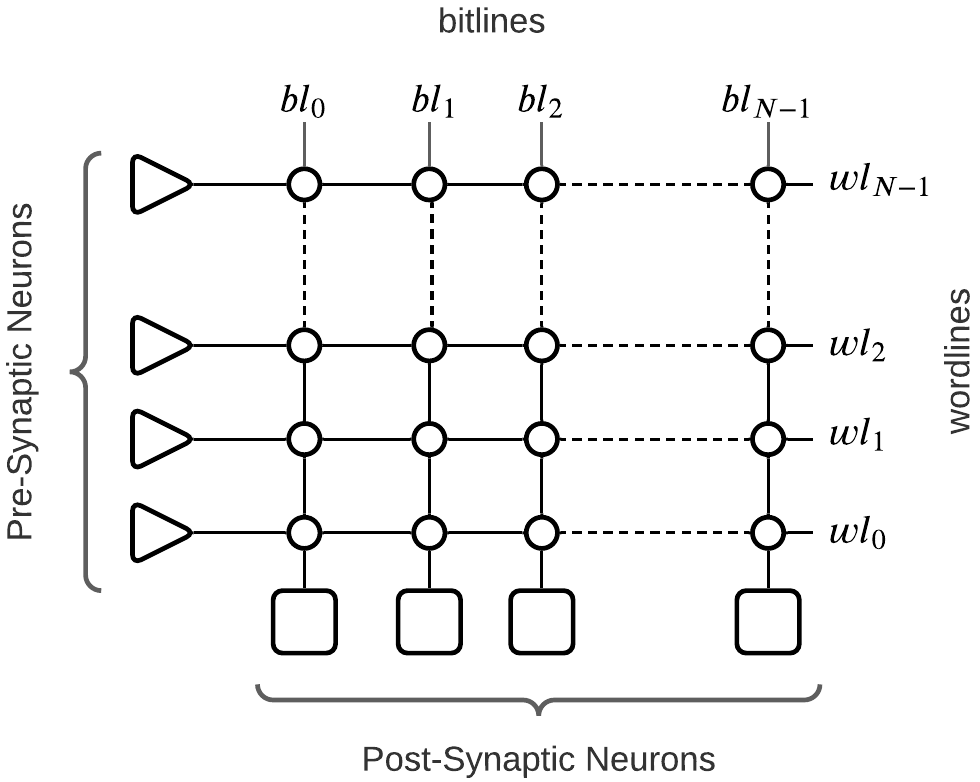} }}%
    \caption{Distributed computing architecture in neuromorphic hardware.}%
    \label{fig:system_architecture}%
\end{figure}

Neuromorphic hardware platforms are used to implement SNN-based machine learning applications~\cite{mead1990neuromorphic}.
%Neuromorphic systems are integrated circuits that mimic the neuro-biological architecture of the central nervous system. Their distributed architecture with in-place neural computation and synaptic storage eliminates the memory bandwidth bottleneck of conventional computers, achieving several orders of magnitude lower energy and significantly higher speed-up in executing SNN-based machine learning applications. 
Table~\ref{tab:hw_examples} shows some of the recently demonstrated neuromorphic hardware platforms with their capacity in terms of number of neurons and synapses.
These platforms are implemented as a many-core hardware~\cite{catthoor2018very} (see Figure~\ref{fig:tile}), where the cores are interconnected via a shared interconnect such as Network-on-Chip~\cite{liu2018neu} and Segmented Bus~\cite{balaji2019exploration}. A neuromorphic core consists of a PE, which implements the neuron circuitry and synaptic cells. A common design practice is to build a PE as an analog crossbar~\cite{liu2015spiking} (see Figure~\ref{fig:crossbar}). In a crossbar, pre-synaptic neuron circuitry acts as current drivers and are placed along each wordline, while post-synaptic neuron circuitry acts as current sinks and are placed along each bitline. Memory cells are placed at the crosspoint of a wordline and bitline, and they store the synaptic weights of an SNN.

\begin{table}[h!]
	\renewcommand{\arraystretch}{0.8}
	\setlength{\tabcolsep}{1pt}
	\caption{Capacity of recent neuromorphic hardware platforms.}
	\label{tab:hw_examples}
	%\vspace{-10pt}
	\centering
	\begin{threeparttable}
	{\fontsize{6}{10}\selectfont
	    %\vspace{-10pt}
		\begin{tabular}{c|cccccccc}
			\hline
			& \textbf{ODIN} & $\mathbf{\mu}$\textbf{Brain} & \textbf{DYNAPs} & \textbf{BrainScaleS} & \textbf{SpiNNaker} & \textbf{Neurogrid} & \textbf{Loihi} & \textbf{TrueNorth}\\
			& \cite{odin} & \cite{mubrain} & \cite{dynapse} & \cite{schemmel2021brainscales} & \cite{spinnaker} & \cite{neurogrid} & \cite{loihi} & \cite{truenorth}\\
			\hline
			\textbf{\# Neurons/core} & 256 & 336 & 256 & 512 & 36K & 65K & 130K & 1M\\
			\textbf{\# Synapses/core} & 64K & 38K & 16K & 128K & 2.8M & 8M & 130M & 256M\\
			\textbf{\# Cores/chip} & 1 & 1 & 1 & 1 & 144 & 128 & 128 & 4096\\
			\hline
			\textbf{\# Chips/board} & 1 & 1 & 4 & 352 & 56 & 16 & 768 & 4096\\
			\hline
			& \multicolumn{8}{|c}{High-performance neuromorphic system}\\
			\textbf{\# Neurons} & 256 & 336 & 1K & 4M & 2.5B & 1M & 100M & 4B\\
			\textbf{\# Synapses} & 256 & 336 & 65K & 1B & 200B & 16B & 100B & 1T\\
			\hline
	\end{tabular}}
	\end{threeparttable}
	%\vspace{12pt}
	%\vspace{-10pt}
\end{table}

A neuromorphic hardware enables distributed and pipelined processing of the operations of an SNN. Additionally, each crossbar in the hardware can implement a maximum of \ineq{N} pre-synaptic neurons per post-synaptic neuron. Therefore, system software frameworks such as NEUTRAMS~\cite{ji2016neutrams}, NeuroXplorer~\cite{neuroxplorer}, Corelet~\cite{corelet}, and PACMAN~\cite{pacman} consist of 1) a compiler, which partitions a SNN model into clusters such that the neurons and synapses of each cluster can be mapped to a crossbar of the hardware, and 2) a run-time manager, which maps the clusters of an SNN to the cores of a many-core hardware. To this end, several mapping strategies are proposed, including optimizing for
energy~\cite{psopart,spinemap,twisha_energy,balaji2020run}, throughput~\cite{sdfsnn,sdfsnn_pp,dfsynthesizer,dfsynthesizer_pp}, resource utilization~\cite{esl20,adarsha_igsc,loihi_mapping,ji2016neutrams}, circuit aging~\cite{ncrtm,balaji2019framework,reneu,song2020case,vts_das}, inference lifetime~\cite{song2021improving}, and endurance~\cite{twisha_endurance,twisha_thermal,espine}.
% resource utilization-aware mapping~\cite{loihi_mapping,esl20,ji2016neutrams}, energy-aware mapping~\cite{psopart,spinemap,twisha_energy}, latency-aware mapping~\cite{das2018dataflow,balaji2019ISVLSIframework}, throughput-aware mapping~\cite{} circuit aging-aware mapping~\cite{reneu,ncrtm,balaji2019framework,song2020case}, endurance-aware mapping~\cite{espine,twisha_endurance}, inference lifetime-aware mapping~\cite{song2021improving}, and thermal-aware mapping~\cite{twisha_thermal}. 
All these mapping techniques use some variants of the SNN partitioning approach proposed in SpiNeMap~\cite{spinemap}.

Recently, dataflow models are used to analyze performance of SNNs implemented on a neuromorphic hardware. There are two strategies proposed in literature -- the \textbf{SDFSNN}~\cite{sdfsnn} and its extended version~\cite{sdfsnn_pp}, which uses dataflow graphs to model an SNN, performing partitioning and mapping explorations with neurons and synapses directly, and the \textbf{DFSynthesizer}~\cite{dfsynthesizer} and its extended version~\cite{dfsynthesizer_pp}, which uses dataflow graphs to only model the clustered SNN, allowing mapping and scheduling of the clusters (a collection of neurons and synapses) to the PEs of a neuromorphic hardware. 

\begin{figure}[h!]
	\centering
	%\vspace{-5pt}
	\centerline{\includegraphics[width=0.89\columnwidth]{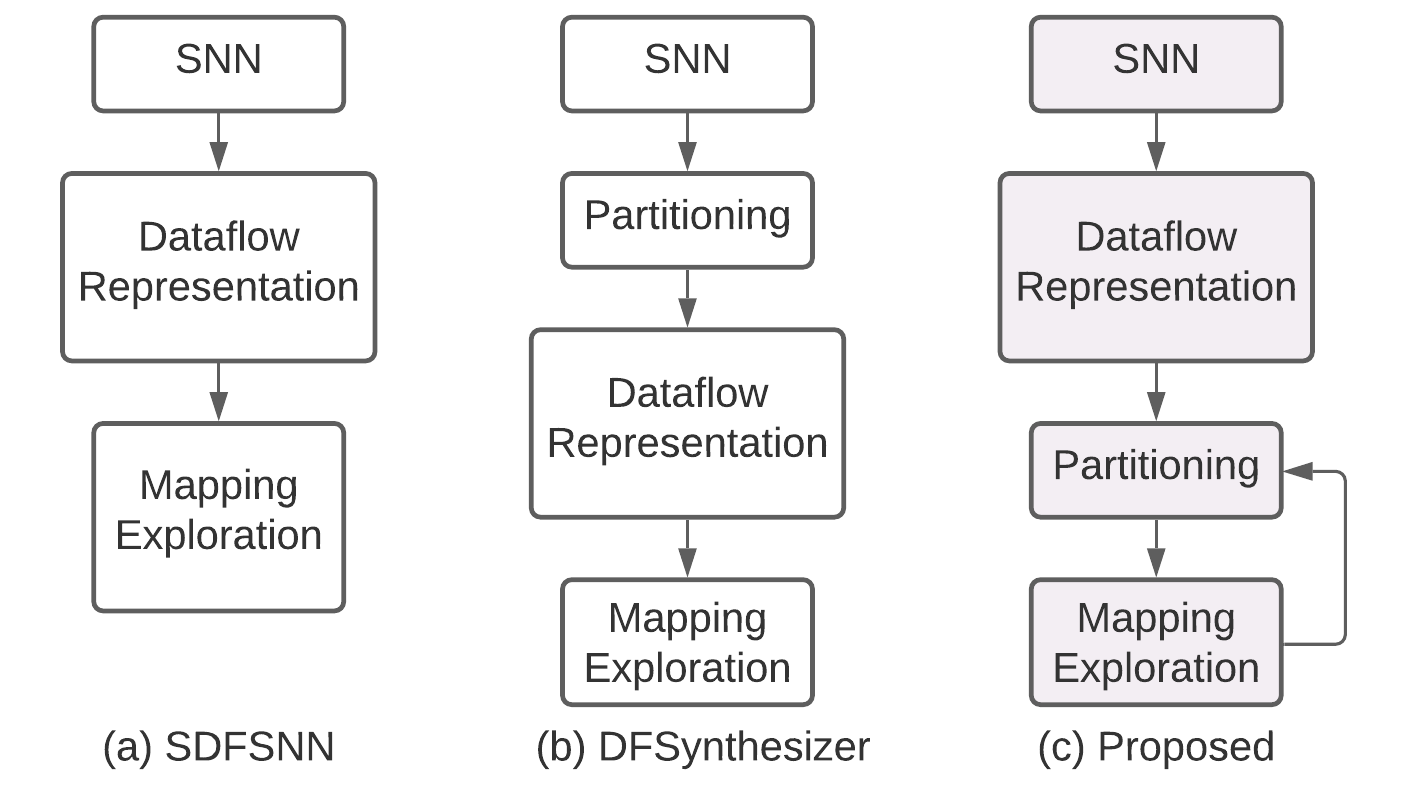}}
	%\vspace{-10pt}
	%\caption{An example of spiking neural network.}
	\caption{Comparing the proposed approach with SDFSNN~\cite{sdfsnn,sdfsnn_pp} and DFSynthesizer~\cite{dfsynthesizer,dfsynthesizer_pp}.}
	%\vspace{-5pt}
	\label{fig:df_difference}
\end{figure}

We show that SDFSNN is not scalable to large SNN models. DFSynthesizer, on the other hand, starts from the clusters of an SNN models and therefore, DFSynthesizer is scalable to large problem sizes. However, we
%does not incorporate all the hardware constraints. We 
show that DFSynthesizer is not able to explore a significant portion of the design space.
Figure~\ref{fig:df_difference} shows at a high-level, how the proposed design flow differs from these existing works. The proposed flow uses an iterative approach involving graph partitioning into clusters followed by cluster mapping to hardware. We describe this flow in details in Section~\ref{sec:design_flow}.

%% file: sections/designflow.tex
\begin{figure}[h!]
	\centering
	%\vspace{-5pt}
	\centerline{\includegraphics[width=0.89\columnwidth]{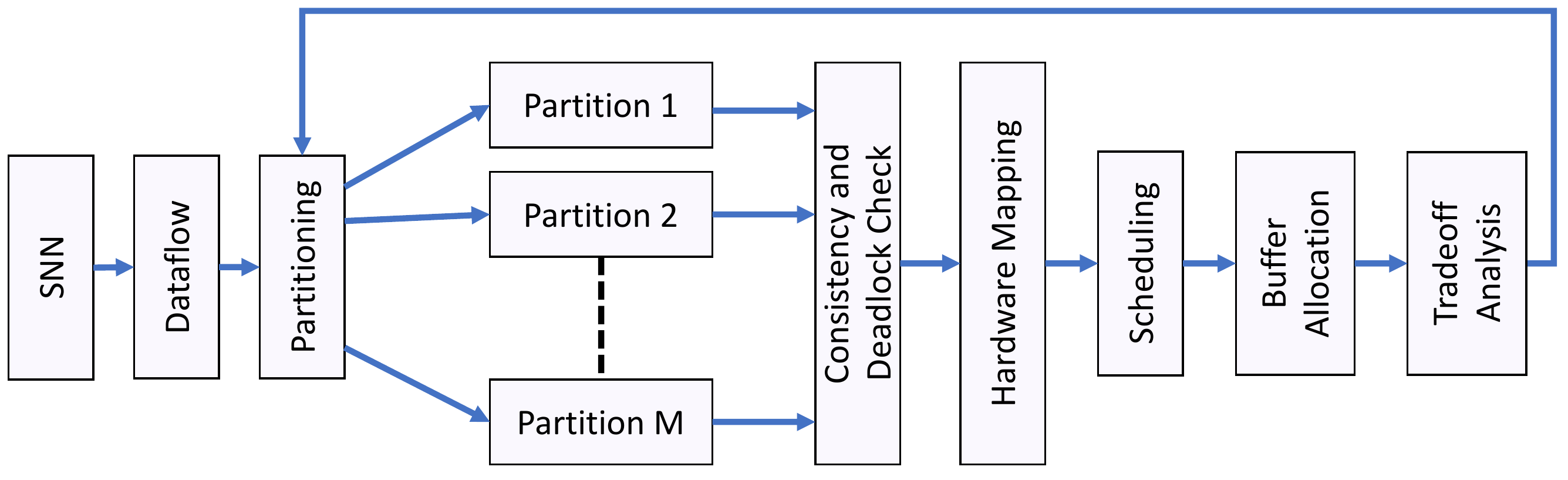}}
	%\vspace{-10pt}
	%\caption{An example of spiking neural network.}
	\caption{Steps of the proposed design flow.}
	%\vspace{-5pt}
	\label{fig:df_full}
\end{figure}

Figure~\ref{fig:df_full} shows the five steps of our design flow. These steps are enumerated below.
\begin{enumerate}
    \item An SNN model is represented using a dataflow graph
    \item The SNN graph is partitioned into clusters using an iterative solution.
    \item Clusters and their connections are analyzed for consistency and deadlock.
    \item The sub-graph representing each cluster is substituted as nodes into the original dataflow graph to generate a dataflow representation of the clustererd SNN.
    \item The clustered SNN graph is mapped to the hardware, where mapping involves allocating a cluster to a core of the hardware.
    \item If more than one clusters are mapped to a core, a list scheduler is used to schedule (order) the execution of these clusters on the core.
    \item A decision is made on the buffer size on each channel. To do so, we make a trade-off between buffer size and throughput of the application.
    \item The design flow explores a new partition and repeats the exploration steps.
\end{enumerate}

% 1) an SNN model is represented using a dataflow graph, 2) the graph is partitioned into clusters using an iterative solution, Third, clusters and their connections are analyzed for consistency and deadlock. Four, the sub-graph representing a cluster is substituted as a node into the original dataflow graph to generate a dataflow representation of the clustererd SNN. Five, this clustered SNN graph is mapped to the hardware, where mapping involves allocating a cluster to a core of the hardware. If more than one clusters are mapped to a core, a list scheduler is used to schedule (order) the execution of these clusters on the core. Six, a decision is made on the storage space assigned to each channel. To do so, we make a trade-off between the used storage space and the throughput of the application. Next, the design flow explores a new partition and the hardware mapping steps are repeated.

%% file: sections/sdfg.tex
We model an SNN as a Synchronous Data Flow Graph (SDFG) for predictable performance analysis~\cite{lee1987synchronous}. SDFGs are commonly used to model streaming applications that are implemented on a multi-core system~\cite{SB00,jiashu2012design,das2018reliable}. 
Both pipelined streaming and cyclic dependencies between tasks can be easily modeled in SDFGs.
These graphs are used to analyze a system in terms of key performance properties such as throughput, execution time, communication bandwidth, and buffer requirement~\cite{Stuijk06dac,singh2013rapiditas,das2012faultRSP}.
Nodes of an SDFG are called \textit{actors}, which
are computed by reading \textit{tokens} from their input ports and writing the results of the computation as tokens on output ports.
Port rates are visualized as annotations on edges. Actor execution is also called \textit{firing}, and it requires a fixed amount of time to execute. Edges in the graph are called \textit{channels} and they represent dependencies among actors.
An actor is said to be {\em ready} when it has sufficient input tokens on all its input channels and sufficient buffer space on all its output channels; an actor can only fire when it is ready.
%A set $Ports$ of ports is assumed., and with each port $p \in Ports$, a finite rate $Rate(p) \in \mathbb{N}\setminus\{0\}$ is associated.
% Formally,
% \begin{Definition}{Actor}
% {An actor $\actor{a}_i$ is a tuple $(I_i,O_i,\tau_i,\mu_i)$ consisting of a set $I_i$ ($\subseteq Ports$) of input ports, a set $O_i$ ($\subseteq Ports$) of output ports with $I_i \cap O_i = \emptyset$, $\tau_i$ is the execution time of $\actor{a}_i$ and $\mu_i$ is its state space, i.e., buffer space needed to communicate tokens on all of its channels.}
% \end{Definition}
% The source of channel $ch_i^j \in C$ is an output port of actor $\actor{a}_i$, the destination is an input port of actor $\actor{a}_j$. All ports of all actors are connected to precisely one channel, and all channels are connected to ports of some actors. The source and the destination port of channel $ch_i^j$ are denoted by $SrcP(ch_i^j)$ and $DstP(ch_i^j)$ respectively. Channels connected to the input and output ports of an actor $\actor{a}_i$ are denoted by $InC(\actor{a}_i)$ and $OutC(\actor{a}_i$) respectively.

%Before an actor $\actor{a}_i$ starts its firing, it requires $Rate(q_i)$ tokens from all $(p,q_i)\in InC(\actor{a}_i)$. When the actor completes execution, it produces $Rate(p_i)$ tokens on every $(p_i,q) \in OutC(\actor{a}_i)$. 
One important property of an SDFG is \textit{throughput}, which is defined as the inverse of its long-term period. A period is the average time needed for one iteration of the SDFG. An iteration is defined as the minimum non-zero execution such that the original state of the SDFG is obtained. This is the performance parameter used in this paper.

\begin{figure}[h!]%
    \centering
    \subfloat[SDFG representation of an SNN.\label{fig:snn_sdfg}]{{\includegraphics[width=4.0cm]{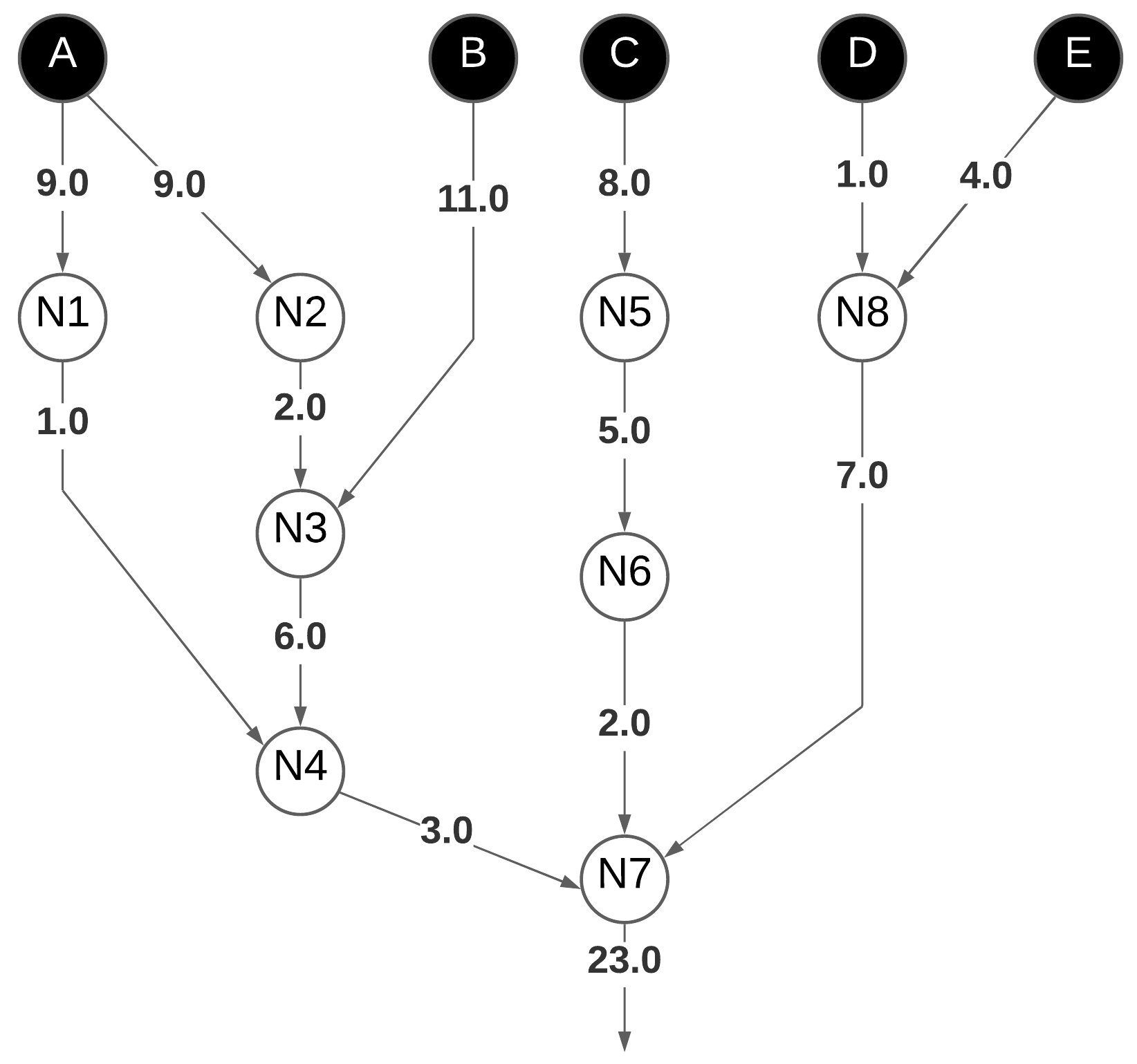} }}%
    \quad
    \subfloat[Partitioned SDFG.\label{fig:snn_cluster_sdfg}]{{\includegraphics[width=4.0cm]{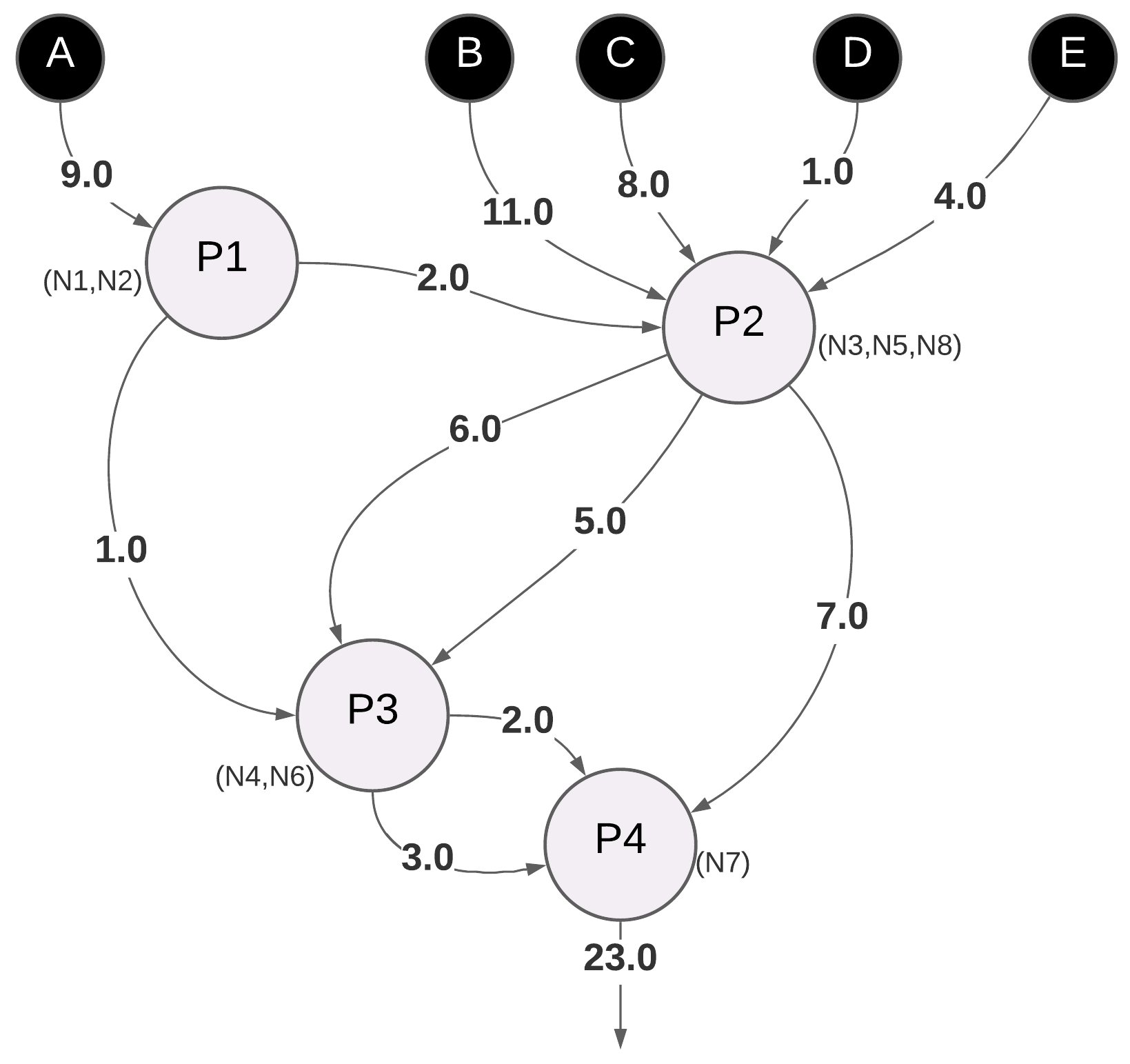} }}%
    \caption{Modeling SNN as a Synchronous Dataflow Graph (SDFG).}%
    \label{fig:sdfg}%
\end{figure}

To model a trained SNN as an SDFG, we consider the average number of spikes per frame on each synaptic connection of the SNN. For image-based applications, which are the primary focus of this work, a frame corresponds to an individual image. For time-series applications such as natural language and bio-signal processing, a frame corresponds to the data collected within a fixed-length timing window. Spike count on synapses of an SNN can be obtained by simulating the trained SNN in a simulator such as Brian~\cite{brian} and PyCARL~\cite{pycarl} using representative training data. Figure~\ref{fig:snn_sdfg} shows an example SNN with 8 neurons (N1-N8) connected to 5 inputs (A-E). Formally,
\begin{Definition}{SNN Graph}
An SNN \ineq{\mathbf{G_{SNN} = (\mathbf{N},\mathbf{S})}} is a directed graph consisting of a finite set \ineq{{\mathbf{N}}} of nodes, representing neurons and a finite set \ineq{{\mathbf{S}}} of edges, representing synapses.%between every pair of pre- and post-synaptic neurons.
%$\tau \in \mathbb{N}\setminus\{0\}$ representing the execution time of a ($\tau(a)$).}
\end{Definition}

Table~\ref{tab:snn_sdfg_mapping} shows the one-to-one mapping of an SNN to SDFG properties. In representing an SNN as an SDFG, we discard the inter-spike interval on synapses, retaining only the spike count. For instance, the neuron N3 (in Fig.~\ref{fig:snn_sdfg}) in our model fires 6 spikes (tokens) at once when it receives 2 spikes from N2 and 11 spikes from input B. 
%This means that the output channel of N3 must have a buffer space of 6 in order for the neuron N3 to fire. 
In practice, however, the 6 output spikes from N3 are generated and transmitted at different times.
%, meaning that a buffer space of 1 is sufficient. In performing the throughput-buffer size trade-offs, we incorporate this reduced requirement of buffer space for SNN dynamics.

\begin{table}[h!]
	\renewcommand{\arraystretch}{0.8}
	\setlength{\tabcolsep}{3pt}
	\caption{One -to-one mapping of SNN to SDFG terminology.}
	\label{tab:snn_sdfg_mapping}
	%\vspace{-10pt}
	\centering
	\begin{threeparttable}
	{\fontsize{6}{10}\selectfont
	    %\vspace{-10pt}
		\begin{tabular}{c|c}
			\hline
			\textbf{SDFG Terminology} & \textbf{SNN Terminology}\\
			\hline
			actor & neuron\\
			channel & synapse \\
			token & spike \\
			\hline
	\end{tabular}}
	\end{threeparttable}
	%\vspace{12pt}
	%\vspace{-10pt}
\end{table}

%Once a SNN model is represented as a SDFG, we use the SDF${}^3$ tool~\cite{sdf3} to analyze its properties. 
Table~\ref{tab:sdfg_properties} reports the average input/output degree and the maximum diameter of the SDFG obtained from the five evaluated machine learning applications (see Section~\ref{sec:results}).

\begin{table}[h!]
	\renewcommand{\arraystretch}{0.8}
	\setlength{\tabcolsep}{3pt}
	\caption{Properties of SDFG representation of evaluated SNN.}
	\label{tab:sdfg_properties}
	%\vspace{-10pt}
	\centering
	\begin{threeparttable}
	{\fontsize{6}{10}\selectfont
	    %\vspace{-10pt}
		\begin{tabular}{l|ccccc}
			\hline
			\multirow{2}{*}{\textbf{Application}} & \multicolumn{2}{|c}{\textbf{In Degree}} & \multicolumn{2}{c}{\textbf{Out Degree}} & \multirow{2}{*}{\textbf{Diameter}}\\
			& \textbf{Max} & \textbf{Average} & \textbf{Max} & \textbf{Average} & \\
			\hline
% 			\textbf{Application} & \textbf{Avg. In Degree} & \textbf{Avg. Out Degree} & \textbf{Max. Depth}\\
% 			\hline
            LeNet & 144 & 73 & 144 & 73 & 4\\
			AlexNet & 102 & 119 & 204 & 119 & 7\\
			ResNet & 288 & 133 & 576 & 134 & 8\\
			DenseNet & 288 & 104 & 576 & 104 & 10\\
			VGG & 288 & 89 & 576 & 89 & 11\\
			\hline
	\end{tabular}}
	\end{threeparttable}
	%\vspace{12pt}
	%\vspace{-10pt}
\end{table}

%% file: sections/partitioning.tex
Each core in a neuromorphic hardware can accommodate only a certain number of pre- and post-synaptic neurons. So, a single core may not be sufficient to map all neurons and synapses of an SNN. In such scenarios where more than one cores are needed, an SNN needs to be partitioned into clusters, where each cluster consists of a subset of neurons and synapses of the original SNN. The partitioning step ensures that a cluster can fit onto a core of the many-core neurommorphic hardware. Spike communication constitutes a significant fraction of the total energy consumption in a neuromorphic hardware~\cite{twisha_energy}. Therefore, SNN partitioning algorithms minimize the spike communication between clusters.
To this end, we propose a novel iterative approach to partitioning an SNN into clusters. Our approach is tightly integrated with cluster mapping explorations to generate better throughput-buffer size trade-off. 

Graph partitioning is an NP-hard problem and has been studied extensively in the context of workload distribution for the efficient use of a distributed memory parallel computer. Several heuristic solutions have been proposed to solve this problem with the objective of minimizing the communication cost between computers and balancing the workload on each computer. A thorough review of these methods and the extensive literature associated with them is beyond the scope of this paper. We chose Kernighan-Lin (KL) recursive graph partitioning approach~\cite{kl}. In the following, we describe how the KL approach is tuned for neuromorphic computing and is integrated inside the proposed iterative solution.

To formulate our partitioning problem, we consider the example of an \ineq{M\times M} analog crossbar, which can accommodate a maximum of \ineq{M} pre-synaptic and \ineq{M} post-synaptic neurons. We represent a partition of the SNN using the binary mapping matrix \ineq{\mathbb{P}\in \mathbb{R}^{|N|\times|C|}}, where \ineq{p_{i,j} = \begin{cases}
1 & \text{if neuron } n_i \text{ is mapped to cluster }c_j\\
0 & \text{otherwise}
\end{cases}
}.

Algorithm~\ref{alg:kl_partitioning} illustrates the pseudo-code for the proposed iterative SNN partitioning. The algorithm runs for \ineq{\eta} iterations, which is an user-defined parameter that controls the design space exploration for throughput-buffer size trade-off. First, the algorithm partitions an SNN graph \ineq{G_{SNN}} by randomly allocating neurons to different clusters (line 2). The total communication cost (measured as the total number of spikes communicated between clusters) is evaluated (line 3). 
By minimizing the total communication cost as formulated, the partitioning algorithm minimizes 1) communication energy, thereby lowering the total energy consumption and 2) congestion on the shared interconnect, thereby reducing the spike latency. Starting from this initial partitioning, the KL approach recursively swaps neurons between clusters, such that the cost is minimized (lines 4-18). To this end, a variable \ineq{\delta} is used to track the reduction of the cost function. \ineq{\delta} is initialized to a very large number (line 4). The algorithm iterates through lines 6 to 17 as long as the improvement in cost is greater than an user defined minimum \ineq{\delta_{min}}. At each iteration, the algorithm performs the following. For each neuron pair (line 6), clusters to which these neurons are mapped are obtained from the partition matrix (lines 7-8). If the two clusters are different (line 9), the partition is changed by swapping the two neurons (line 10). If this new change is valid (i.e., both the clusters satisfy the hardware constraint), then the new cost is evaluated (lines 11-12). If the cost is lower than the initial cost (line 13), the neuron swap is made permanent and the reduction in the cost function is evaluated (lines 14-16). The KL partitioning terminates by generating a clustered SNN graph \ineq{G_{CSNN}} from the partition matrix by replacing the subgraph of each cluster as a node (line 19). Figure~\ref{fig:snn_cluster_sdfg} shows the clusters generated from the original SNN of Figure~\ref{fig:snn_sdfg}.
A clustered SNN graph is formally defined as
\begin{Definition}{Clustered SNN Graph}
A clustered SNN graph \ineq{\mathbf{G_{CSNN} = (\mathbf{C},\mathbf{E})}} is a directed graph consisting of a finite set \ineq{{\mathbf{C}}} of clusters and a finite set \ineq{{\mathbf{E}}} of edges between these clusters.
%$\tau \in \mathbb{N}\setminus\{0\}$ representing the execution time of a ($\tau(a)$).}
\end{Definition}

\begin{algorithm}[h]
	\scriptsize{
 		\KwIn{\ineq{G_{SNN}= (N,S)}}
 		%\KwOut{Clustered SNN \ineq{G_{CSNN}= (C,E)}}
 		\For(\tcc*[f]{Run for $\eta$ iterations}){$r=0;r<\eta;r\texttt{++}$}{
 		    $\mathbb{P}^{init} = $\texttt{InitPartition()}\tcc*[r]{Initial partition}
 		    Evaluate $Cost^{init}$\tcc*[r]{Evaluate comm. cost of this initial partition}
 		    \tcc*[l]{KL Partitioning begins here}
 		    $\delta = \infty$ \tcc*[r]{Set a large value to $\delta$}
 		    \While(\tcc*[f]{Repeat until the improvement in cost is not significant}){$\delta > \delta_{min}$} {
 		    \For(\tcc*[f]{For each pair of nodes in the SDFG $G_{SNN}$}){$n_i,n_j\in N$}{
 		        $k = $\texttt{argmax }$\mathbb{P}^{init}(i,:)$ \tcc*[r]{Find the cluster where neuron $n_i$ is mapped}
 		        $l = $\texttt{argmax }$\mathbb{P}^{init}(j,:)$ \tcc*[r]{Find the cluster where neuron $n_j$ is mapped}
 		        \uIf(\tcc*[f]{If the cluster of $n_i$ and $n_j$ are different}){$k \neq l$}{
 		            $\mathbb{P}^{new} = \mathbb{P}^{init}|p_{i,k} = 0,p_{j,l} = 0, p_{i,l} = 1, p_{j,k} = 1$ \tcc*[r]{Swap the neurons $n_i$ and $n_j$}
 		            \uIf(\tcc*[f]{If the neuron swap is valid}){$\sum_u \mathbb{P}^{new}(u,k) \leq M$ and $\sum_v \mathbb{P}^{new}(v,l) \leq M$}{
 		                Evaluate $Cost^{new}$ \tcc*[r]{Evaluate comm. cost of this new partition}
 		                \uIf(\tcc*[f]{If the cost reduces}){$cost^{new} <  cost^{init}$}{
 		                    $\mathbb{P}^{init} = \mathbb{P}^{new}$\tcc*[r]{Retain the swap}
 		                    $\delta = cost^{init} - cost^{new}$\tcc*[r]{Retain the improvement in cost}
 		                    $cost^{init} = cost^{new}$\tcc*[r]{Set new cost}
 		                 }
 		            }
 		        }
 		    }
 		    }
 		    Generate $G_{CSNN} = (C,E)$ using $\mathbb{P}^{init}$ \tcc*[r]{Generate the clustered graph from the mapping}
 		\tcc*[l]{KL Partitioning ends here}
 		\texttt{Mapping}($G_{CSNN}$)\tcc*[r]{Perform hardware mapping}
 		%\texttt{Scheduling}($G_{CSNN}$)\tcc*[r]{Schedule the clusters using a list-scheduler}
 		%Evaluate throughput-buffer tradeoffs
 		}
 	}
	\caption{Partitioning SDFG graph $G_{SNN}$.}
	\label{alg:kl_partitioning}
\end{algorithm}

The partitioning algorithm uses the clustered SNN graph to perform hardware mapping (line 20) for throughput-buffer size trade-off (line 21). Finally, the algorithm is repeated to explore a new design space, starting from another initial partitioning (line 2). We next describe this hardware mapping.

%% file: sections/mapping.tex
In order to perform the hardware mapping exploration of a clustered SNN graph, we represent a many-core neuromorphic hardware using the hardware graph defined as
\begin{Definition}{Neuromorphic Hardware Graph}
A neuromorphic hardware graph \ineq{\mathbf{H = (\mathbf{T},\mathbf{L})}} is a directed graph consisting of a finite set \ineq{{\mathbf{T}}} of cores and a finite set \ineq{{\mathbf{L}}} of links between these cores.
%$\tau \in \mathbb{N}\setminus\{0\}$ representing the execution time of a ($\tau(a)$).}
\end{Definition}

\begin{Definition}{Core and Link}
{A core $\actor{t}_i$ is a tuple $\langle I_i,O_i,\tau_i,inC(i),outC(i),inB(i),outB(i)\rangle$ consisting of a set $I_i$ ($\subseteq Ports$) of input ports, a set $O_i$ ($\subseteq Ports$) of output ports with $I_i \cap O_i = \emptyset$, $\tau_i$ is the execution time of $\actor{t}_i$, $\left(inC(i),outC(i)\right)$ is the maximum number of incoming and outgoing connections supported by $\actor{t}_i$, and $\left(inC(i),outC(i)\right)$ is its maximum incoming and outgoing bandwidth. Each link \ineq{l_{i,j}\in \mathbf{L}} connecting cores \ineq{\actor{t}_i} and \ineq{\actor{t}_j} is associated with a latency \ineq{t_{i,j}}, which is the time it takes to communicate a spike packet on this link.}
\end{Definition}

\begin{figure}[h!]
	\centering
	%\vspace{-5pt}
	\centerline{\includegraphics[width=0.99\columnwidth]{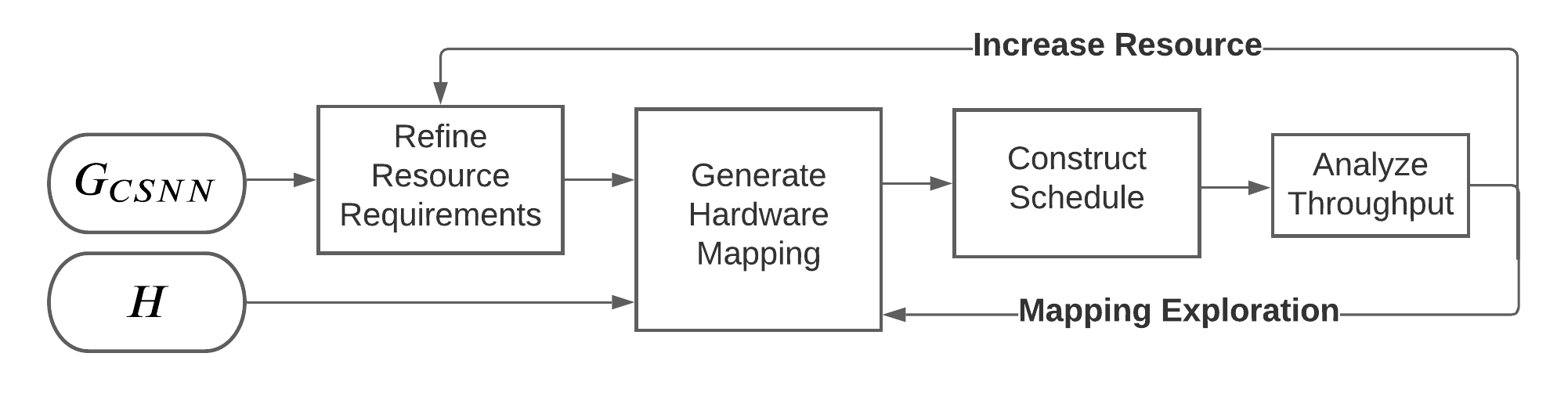}}
	%\vspace{-10pt}
	%\caption{An example of spiking neural network.}
	\caption{Design space exploration to perform throughput-buffer size trade-off.}
	%\vspace{-5pt}
	\label{fig:dse}
\end{figure}

Figure~\ref{fig:dse} shows our design-space exploration framework for mapping an SNN to a many-core neuromorphic hardware. The flow starts with refining resource requirements of a clustered SNN graph. An application graph specifies only the resource requirement of its clusters. Estimating resource requirements of its edges (i.e., buffer size and bandwidth) is performed in this first step of the flow.
In the next step, the flow maps each cluster to a core. For this mapping, a static-order schedule is constructed for each core that maps more than one clusters. Next, the throughput is computed and the exploration is continued starting with a different cluster-to-core mapping. Finally, the flow iterates back to step 1 and increases the buffer size assigned to edges in order to explore a new design space.

\begin{figure}[h!]
	\centering
	%\vspace{-5pt}
	\centerline{\includegraphics[width=0.99\columnwidth]{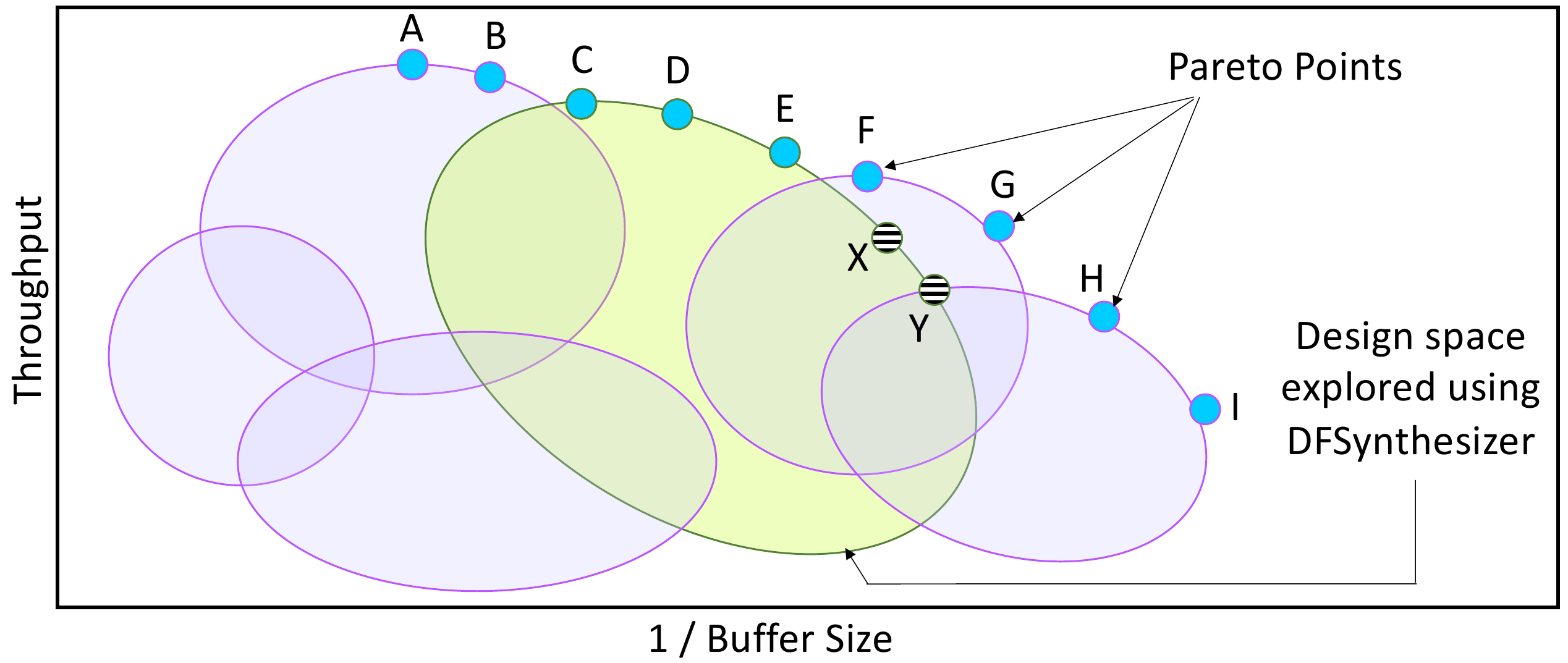}}
	%\vspace{-10pt}
	%\caption{An example of spiking neural network.}
	\caption{Demonstration of design space exploration of throughput and buffer size, and the selection of Pareto points.}
	%\vspace{-5pt}
	\label{fig:pareto_demo}
\end{figure}

Figure~\ref{fig:pareto_demo} illustrates the selection of Pareto points using our design space exploration approach. There are 9 Pareto points (A-I) obtained from 6 design spaces (big circles), which correspond to 6 distinct partitioning of an SNN. The design space using DFSynthesizer is shown in the figure with a different color circle. We observe that Pareto points C, D, and E are common to both DFSynthesizer and the proposed approach. However, Pareto points X and Y of DFSynthesizer are discarded in favor of better solutions (Pareto points F, G, and H) obtained using the proposed approach. We conclude that the proposed approach can generate better trade-off between throughput and buffer size.

\subsection{Refining Resource Requirements}
%The graph generated after partitioning an SNN may contain cyclic dependency. Consistency and absence of deadlock 
Spikes that are communicated on edges of a clustered SNN graph must be stored in a buffer. The amount of buffer that is allocated to these edges has a large impact on the achieved throughput of an application. Allocating more buffer to an edge might increase the throughput because it may increase pipelining opportunities.
Typically, buffer size is chosen such that the throughput requirement is met~\cite{ade1997data,geilen2005minimising,govindarajan2002minimizing}. However, the throughput requirement is not known beforehand. Therefore, a trade-off must be made between the realizable throughput and the buffer allocated to the edges of the clustered SNN graph.

% The storage space of an edge is in principle, unbounded, i.e., it can contain arbitrary number of spikes. In practice, however, the storage space is bounded. 

We use the \texttt{SDF}${}^3$ tool~\cite{sdf3} to perform this throughput-buffer size trade-offs, i.e., generating different \emph{Pareto points}. \texttt{SDF}${}^3$ uses a fast technique involving the construction of abstract dependency graph from the clustered SNN graph to estimate the maximum throughput for a given buffer size by considering its mapping to a single-core neuromorphic hardware~\cite{stuijk2006exploring}. This simplifies the analysis in the absence of hardware mapping information, which is obtained in the subsequent steps. However, to make the analysis relevant for multi-core neuromorphic hardware, we consider separate buffer on each edge, with the total buffer size obtained by adding the buffer sizes allocated to different edges~\cite{stuijk2006exploring}.

\begin{figure}[h!]
	\centering
	%\vspace{-5pt}
	\centerline{\includegraphics[width=0.99\columnwidth]{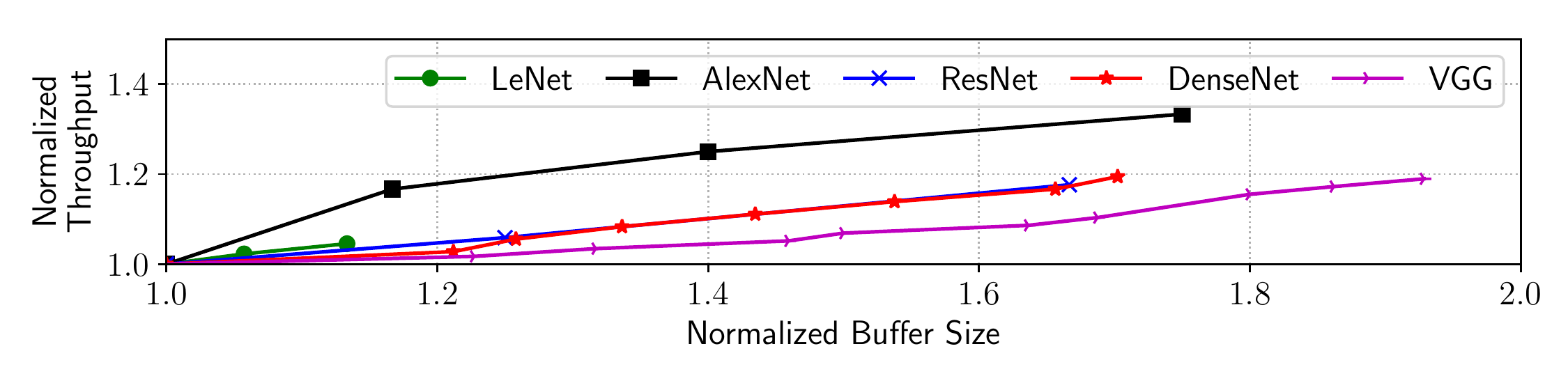}}
	%\vspace{-10pt}
	%\caption{An example of spiking neural network.}
	\caption{Throughput buffer size tradeoffs.}
	%\vspace{-5pt}
	\label{fig:trade_offs}
\end{figure}

Figure~\ref{fig:trade_offs} reports the Pareto points for the five evaluated applications. We observe that throughput of these applications increases with an increase in the allocated buffer size. This is because with more buffers on edges, clusters can be executed earlier whenever tokens are ready, which increases throughput.

\subsection{Generating Hardware Mapping}
For each Pareto point, a hardware mapping exploration is performed, where mapping involves placing each cluster of the clustered SNN graph on to a core of the hardware. To this end, we use an instance of the Particle Swarm Optimization (PSO)~\cite{pso}, a meta-heuristic algorithm used to search for the optimum solution of an optimization problem. The mapping problem is indicated using the matrix \ineq{\mathcal{M}\in \mathbb{R}^{|C|\times|T|}}, where \ineq{m_{i,j} = \begin{cases}
1 & \text{if cluster } c_i \text{ is mapped to core }t_j\\
0 & \text{otherwise}
\end{cases}
}.

The mapping constraint is the following:
\begin{itemize}
    \item A cluster can be mapped to only one crossbar, i.e.,
    \begin{equation}
        \label{eq:mapping_constraint_1}
        \footnotesize \sum_j m_{ij} = 1~~~\forall i
    \end{equation}
\end{itemize}

The optimization problem is to maximize the throughput of an SNN represented as an SDFG. For computing throughput, we use the \texttt{SDF}${}^3$ tool, which estimates throughput of an SDFG based on its self-timed execution~\cite{ghamarian2006throughput}. To do so, we integrate the tool inside our PSO formulation, allowing to estimate the throughput for a given allocation of clusters to cores. Therefore, the fitness function is represented as \ineq{{F} = \text{SDF}^3(\mathcal{M})}.

For PSO, we instantiate \ineq{n_p} swarm particles. The position of these particles are solutions to the fitness function, and they represent different cluster-to-core mappings. Each particle also has a velocity with which it moves in the search space to find the optimum solution. During the movement, a particle updates its position and velocity according to its own experience (closeness to the optimum) and also experience of its neighbors. We introduce the following notations.
 
\begin{footnotesize}
 	\begin{align}
 	\label{eq:pso_defn}
 	D = |{C}|\times|{T}| &= \text{dimensions of the search space}\\
 	\mathbf{\Theta} = \{\mathbf{\theta}_l\in\mathbb{R}^{D}\}_{l=0}^{n_p-1} &= \text{positions of particles in the swarm}\nonumber \\
 	\mathbf{V} = \{\mathbf{v}_l\in\mathbb{R}^{D}\}_{l=0}^{n_p-1} &= \text{velocity of particles in the swarm}\nonumber 
 	\end{align}
 \end{footnotesize}
Position and velocity of swarm particles are updated, and the fitness function is computed as

\begin{footnotesize}
	\begin{align}
	\label{eq:pos_vel_update}
	\mathbf{\Theta}(t+1) &= \mathbf{\Theta}(t) + \mathbf{V}(t+1)\\
	\mathbf{V}(t+1) &= \mathbf{V}(t) + \varphi_1\cdot\Big(P_{\text{best}}-\mathbf{\Theta}(t)\Big) + \varphi_2\cdot\Big(G_{\text{best}}-\mathbf{\Theta}(t)\Big)\nonumber\\
	F(\theta_l) &= \text{SDF}^3(M_l)\nonumber
	\end{align}
\end{footnotesize}
\normalsize where $t$ is the iteration number, $\varphi_1,\varphi_2$ are constants and $P_{\text{best}}$ (and $G_{\text{best}}$) is the particles own (and neighbors) experience. 
Finally, local and global bests are updated as

\begin{footnotesize}
    \begin{align}
    \label{eq:pbest}
        && P_\text{best}^l = F(\theta_l) \text{ if } F(\theta_l) < F(P_\text{best}^l)\nonumber \\
        && G_\text{best} = \displaystyle \min_{l=0,\dots n_p-1} P_\text{best}^l
    \end{align}
\end{footnotesize}

The mapping with the highest throughput is retained.

\subsection{Constructing Static-Order Schedule and Estimating Throughput}
To estimate throughput, the \texttt{SDF}${}^3$ tool constructs a static order schedule for each core of the neuromorphic hardware. This is to arbitrate the access of shared resources of a core (input/output channel, synaptic memory, etc.) among neurons mapped to the core. 
%To this end, we note that the clustered SNN with its hardware mapping (i.e., a position of PSO particles) is an SDFG. Therefore, it can be scheduled using the Earliest Deadline First (EDF) algorithm~\cite{singh2017applying}. %Once schedules are constructed for every core of the hardware, throughput is computed as follows.  
A list-scheduler is used to construct these static-order schedules for all cores at once. The schedules are constructed via an execution of the clustered SNN graph mapped to the cores of hardware, assuming that for each core 50\% of the available time wheel is allocated to the SNN graph. The latency to communicate spikes between cores is taken into account in the schedule construction. When an neuron becomes ready, it does not start its firing immediately. Instead the neuron is added to the ready list of the core it is bound to. When no neuron is firing on the core, the first ready neuron is removed from the list and its firing is started. The neuron ends firing after the time it takes to generate a spike. At this moment, the neuron is added to the schedule of the core. The execution ends as soon as a recurrent state is found. At this point, a finite-length schedule has been constructed for each core. After constructing the schedule, an optimization is performed to remove all recurrent occurrences of the same scheduling sequence. The static-order schedule on each core consists of a transient phase followed by a steady-state phase~\cite{DasKV14}.
Throughput is computed as the inverse of the long-term period in the steady-state.

%% file: sections/evaluation.tex
We conduct all simulations on a Lambda workstation, which has AMD Threadripper 3960X with 24 cores, 128 MB cache, 128 GB RAM, and 2 RTX3090 GPUs. We evaluate 5 convolutional neural network (CNN) models -- LeNet, AlexNet, ResNet (ResNet18), DenseNet, and VGG (VGG16). All these models trained on the CIFAR-10 dataset. We use Keras~\cite{keras} to train these models. Trained models are converted to SNN using the conversion toolbox~\cite{jolpe18,dfsynthesizer_pp} and simulated using PyCARL~\cite{pycarl} with the CARLsim backend simulator~\cite{carlsim}. All spiking neurons are programmed as integrate-and-fire (IF) type~\cite{ifneuron}. The simulator is configured to use OxRRAM NVM model as the synaptic cell~\cite{mallik2017design}.

Our hardware simulation framework includes a cycle-level multi-core neuromorphic system simulator~\cite{neuroxplorer}. We configure this framework to simulate Loihi neuromorphic PEs with parameters listed in Table~\ref{tab:hw_parameters}.

%\vspace{-10pt}
\begin{table}[h!]
    \caption{Major simulation parameters extracted from Loihi~\cite{loihi}.}
	\label{tab:hw_parameters}
	%\vspace{-5pt}
	\centering
	{\fontsize{8}{10}\selectfont
		\begin{tabular}{lp{6cm}}
			\hline
			Neuron technology & 16nm CMOS (original design is at 14nm FinFET)\\
			\hline
			Synapse technology & {HfO${}_2$-based OxRRAM}~\cite{mallik2017design}\\
			\hline
			Supply voltage & 1.0V\\
			\hline
			Energy per spike & 23.6pJ at 30Hz spike frequency\\
			\hline
			Energy per routing & 3pJ\\
			\hline
			Switch bandwidth & 3.44 G. Events/s\\
			\hline
% 			NVM related & 1T-1R\\
% 			& PCM cell SET: 24 cycles\\
% 			& PCM cell RESET: 18 cycles\\
% 			& Program \& Verify: 35 cycles\\
% 			\hline
	\end{tabular}}
\end{table}

%% file: sections/results.tex
\subsection{Maximum Throughput}\label{sec:throughput_results}
\begin{figure}[h!]
	\centering
	%\vspace{-5pt}
	\centerline{\includegraphics[width=0.99\columnwidth]{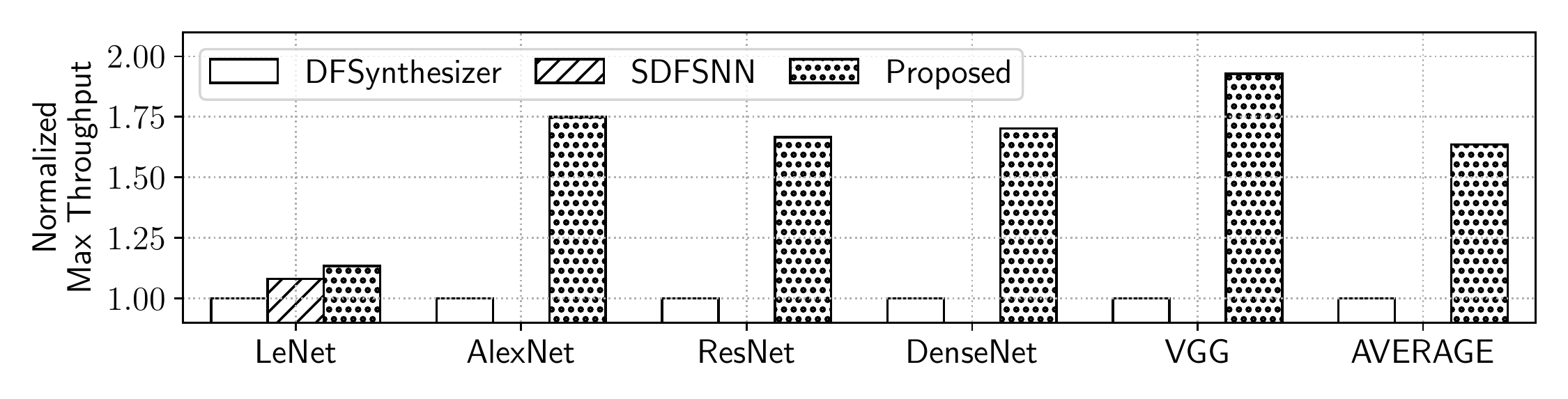}}
	%\vspace{-10pt}
	%\caption{An example of spiking neural network.}
	\caption{Maximum throughput.}
	%\vspace{-5pt}
	\label{fig:max_thr}
\end{figure}

Figure~\ref{fig:max_thr} reports the maximum throughput obtained using the proposed design-flow compared to DFSynthesizer and SDFSNN for the five CNN applications. Results are normalized to DFSynthesizer. We make the following two key observations. 

First, the maximum throughput of SDFSNN is 8\% higher than DFSynthesizer. This is because, SDFSNN performs throughput analysis treating an entire SNN graph as an SDFG, and performing both partitioning and hardware placement at once during its analysis stage. DFSynthesizer, on the other hand, applies dataflow analysis technique on an SNN model that is already partitioned into clusters. Therefore, the search space of DFSynthesizer is smaller than SDFSNN (see Figure~\ref{fig:pareto_demo}), resulting in lower maximum throughput. However, SDFSNN is not scalable for large problem sizes due to its integrated partitioning and placement steps. For these applications, SDFSNN fails to generate a mapping solution as we see in the figure. Second, maximum throughput of the proposed design flow is the highest for all CNN models. The maximum throughput is on average 63\% higher than DFSynthesizer for all CNN models, and 5\% higher than SDFSNN for the LeNet model. The improvement is because the proposed design flow uses iterative approach, performing partitioning using the KL heuristic and throughput analysis exploiting the rich semantics of SDFG. Due to the use of KL heuristic, the proposed design flow is scalable to large problem sizes. Additionally, due to creating different partitioning alternatives and performing design-space exploration with them, the proposed design flow is able to explore a much larger throughput-buffer size search space than DFSynthesizer.  

\subsection{Buffer Size}\label{sec:buffer_results}
\begin{figure}[h!]
	\centering
	%\vspace{-5pt}
	\centerline{\includegraphics[width=0.99\columnwidth]{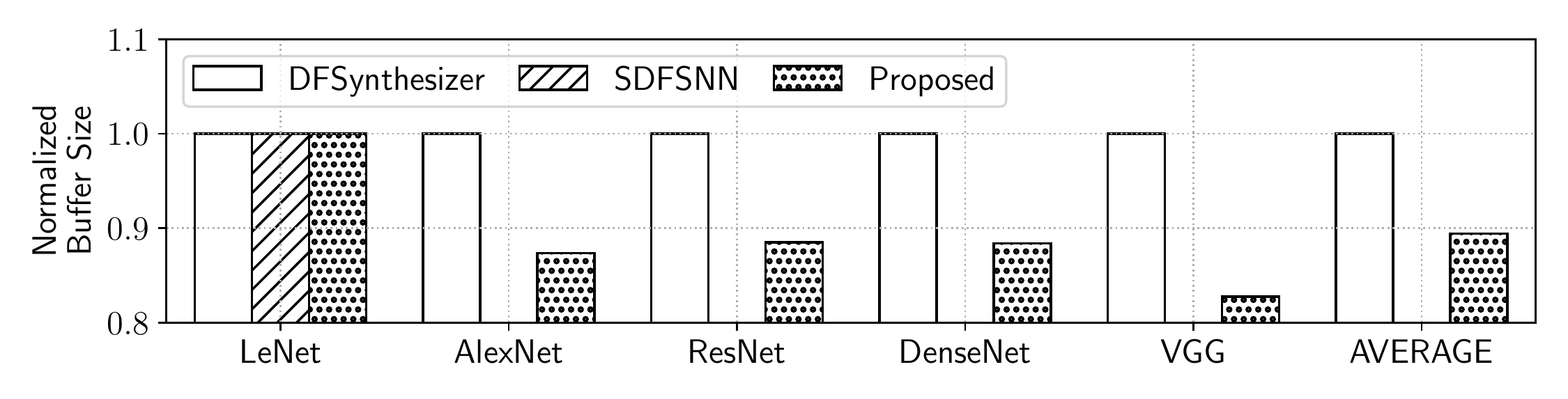}}
	%\vspace{-10pt}
	%\caption{An example of spiking neural network.}
	\caption{Minimum buffer size.}
	%\vspace{-5pt}
	\label{fig:buff_size}
\end{figure}

Figure~\ref{fig:buff_size} reports the minimum buffer size needed to achieve a throughput constraint for each evaluated model using the three approaches. The throughput constraint is set to 70\% of the highest throughput obtained using the proposed design flow. We selected this throughput constraint as a case study because both DFSynthesizer and SDFSNN are not able to find a mapping solution for throughput constraint set to anything higher than this value due to limited size of their exploration space (see Section~\ref{sec:search_space}). Results for each application are normalized to DFSynthesizer. We make the following three key observations.

First, the minimum buffer size needed to achieve the throughput constraint is the least for the proposed design flow (on average 10\% lower than DFSynthesizer). This is because the proposed design flow is able to explore larger design space than DFSynthesizer, which we have discussed in Section~\ref{sec:throughput_results} (see also Figure~\ref{fig:pareto_demo}). Second, the buffer size needed for LeNet in order to achieve the throughput constraint is the same for all three approaches. Combining results of buffer size and maximum throughput for LeNet, we conclude that for a given amount of buffer size, the proposed design flow results in higher throughput than the two state-of-the-art dataflow based mapping frameworks.

\begin{figure}[h!]
	\centering
	%\vspace{-5pt}
	\centerline{\includegraphics[width=0.99\columnwidth]{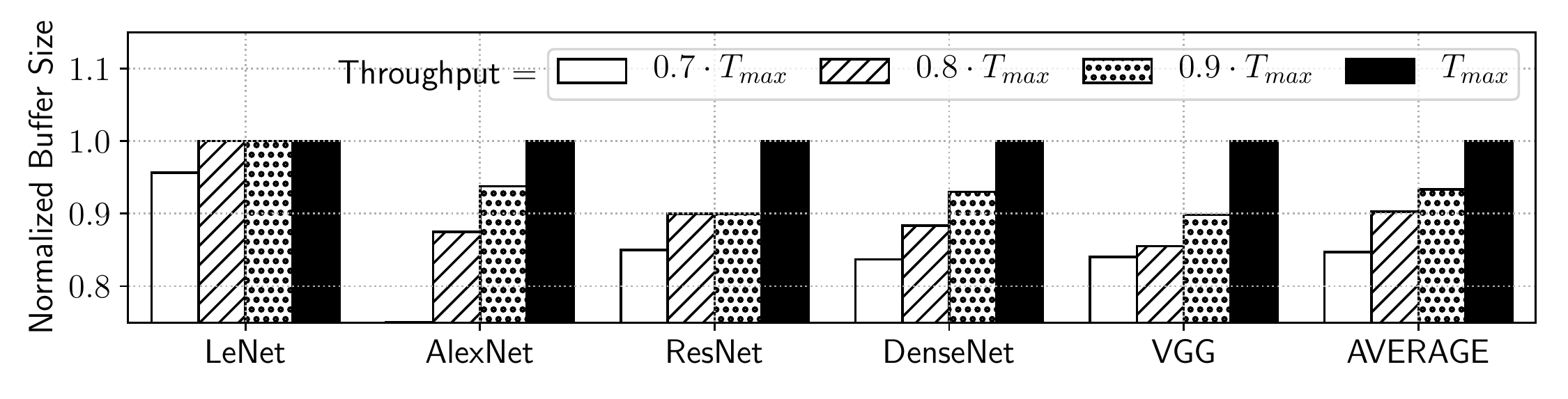}}
	%\vspace{-10pt}
	%\caption{An example of spiking neural network.}
	\caption{Minimum buffer size for different throughput constraints.}
	%\vspace{-5pt}
	\label{fig:buff}
\end{figure}

To give further insight, Figure~\ref{fig:buff} reports the minimum buffer size needed to achieve different throughput constraints using the proposed design flow. There are four settings evaluated for each application -- minimum buffer size needed to achieve 70\%, 80\%, 90\%, and 100\% of the highest throughput (\ineq{T_{max}}).
%the throughput constraint (\ineq{T_c}), that to achieve 25\% and 50\% higher than the throughput constraint, and the buffer size to to achieve the maximum throughput \ineq{T_{max}}. Here, throughput constraint \ineq{T_c} is set to 75\% of \ineq{T_{max}}. 
Results for each application are normalized to the buffer size needed to achieve the highest throughput. We make the following two key observations.

First, to achieve 70\%, 80\%, and 90\% of the highest throughput, the minimum buffer size needed in the proposed design flow is on average 15.3\%, 9.7\%, and 6.7\% lower than the buffer size needed to achieve the highest throughput. These results show that for scalable throughput application, i.e., those applications with acceptable throughput degradation, the buffer requirement of the hardware can be reduced significantly using the proposed design flow. Second, for LeNet, which is a smaller CNN model compared to the rest, there are only a fewer Pareto points generated using the design flow. Therefore, we see no change in the minimum buffer size for this application as we increase the throughput requirement from 80\% to 100\% of the highest throughput.

\subsection{Design Space Explorations}\label{sec:search_space}
Table~\ref{tab:dse} reports performance of the design-space exploration using the proposed design flow compared to DFSynthesizer. For the proposed design flow, we report results for three different settings of the user-defined parameter \ineq{\eta}. Results in Sections~\ref{sec:throughput_results} and \ref{sec:buffer_results} are obtained by setting \ineq{\eta = 10}. Design-space exploration is compared in terms of the number of Pareto points generated during the exploration and the time (s) it takes to explore the design space. We make the following four key observations.

\begin{table}[h!]
	\renewcommand{\arraystretch}{0.8}
	\setlength{\tabcolsep}{2pt}
	\caption{Design Space Exploration.}
	\label{tab:dse}
	%\vspace{-10pt}
	\centering
	\begin{threeparttable}
	{\fontsize{6}{10}\selectfont
	    %\vspace{-10pt}
		\begin{tabular}{c|cc|cccccc}
			\hline
			\multirow{3}{*}{\textbf{Model}} & \multicolumn{2}{|c}{\textbf{DFSynthesizer}} & \multicolumn{2}{|c}{$\mathbf{\eta = 1}$} & \multicolumn{2}{|c}{$\mathbf{\eta = 5}$} & \multicolumn{2}{|c}{$\mathbf{\eta = 10}$}\\ \cline{2-9}
			& \textbf{Pareto} & \textbf{Exploration} & \textbf{Pareto} & \textbf{Exploration} & \textbf{Pareto} & \textbf{Exploration} & \textbf{Pareto} & \textbf{Exploration} \\
			& \textbf{Points} & \textbf{Time (s)} & \textbf{Points} & \textbf{Time (s)} & \textbf{Points} & \textbf{Time (s)} & \textbf{Points} & \textbf{Time (s)}\\
			\hline
			LeNet & 4 & 108 & 4 & 432 & 4 & 3024 & 4 & 6288\\
            AlexNet & 4 & 1463 & 7 & 4389 & 9 & 39501 & 10 & 75519\\
            ResNet & 5 & 2723 & 8 & 10892 & 9 & 65352 & 11 & 118872\\
            DenseNet & 4 & 4399 & 8 & 8798 & 12 & 70384 & 13 & 176144\\
            VGG & 6 & 6563 & 7 & 13126 & 10 & 131260 & 12 & 293940\\
			\hline
	\end{tabular}}
	\end{threeparttable}
	%\vspace{12pt}
	%\vspace{-10pt}
\end{table}

First, the number of Pareto points obtained using the proposed design flow is higher than DFSynthesizer. For smaller models such as LeNet, the number of Pareto points are comparable. However, for larger models, the proposed design flow generates higher number of Pareto points than DFSynthesizer, resulting in higher maximum throughput (Section~\ref{sec:throughput_results}) and lower buffer requirement (Section~\ref{sec:buffer_results}).
Second, the number of Pareto points increases with increase in \ineq{\eta}. This is because with more iterations of the partitioning algorithm (Algorithm~\ref{alg:kl_partitioning}), the proposed design flow can explore larger design space, leading to generating more Pareto points.
Third, the exploration time using DFSynthesizer is the least. This is because, DFSynthesizer's design space exploration is limited to exploration using the clusters only, which are fewer than the number of neurons. The proposed design flow explores the design space using neurons. Therefore, the exploration time is higher than DFSynthesizer, even with \ineq{\eta = 1}.
Finally, the exploration time of the proposed design flow increases with increase in \ineq{\eta}. Designer can select \ineq{\eta} based on the required throughput-buffer size tradeoff.

%% file: sections/conclusions.tex
We propose a design flow for predictable mapping of SNN-based machine learning models to many-core neuromorphic hardware. The design flow consists of an iterative approach to partition an SNN into clusters such that each cluster can be mapped to a core of the many-core hardware. The partitioning step minimizes the inter-cluster spike communication, which improves latency. The design flow then uses an instance of the Particle Swarm optimization (PSO) to generate SNN mapping solutions, exploring the design space between throughput and buffer size requirement of the cores. Pareto optimal mappings are provided to system designer. We evaluate our design flow using large-scale spiking CNN models. Results demonstrate 63\% higher maximum throughput and 10\% lower buffer requirement than state of the art mapping solutions.